\documentclass[10pt, a4paper, conference, compsocconf]{IEEEtran}
\usepackage{times}
\usepackage{epsfig}
\usepackage{graphicx}
\usepackage{amsmath}
\usepackage{amssymb}
\usepackage{multirow}
\usepackage[pagebackref=true,breaklinks=true,letterpaper=true,colorlinks,bookmarks=false]{hyperref}
\ifCLASSINFOpdf
\else
\fi
\begin{document}
\title{UBSegNet: Unified Biometric Region of Interest Segmentation Network}


\author{\IEEEauthorblockN{Ranjeet Ranjan Jha, Daksh Thapar, Aditya Nigam}
\IEEEauthorblockA{ School of Computing and Electrical Engineering, \\
Indian Institute of Technology Mandi\\
Mandi, India - 175005\\
Email: d16044,s16007@students.iitmandi.ac.in, aditya@iitmandi.ac.in}
\and
\IEEEauthorblockN{Shreyas Malakarjun Patil}
\IEEEauthorblockA{ Department of Electrical Engineering, \\
 Indian Institute of Technology, Jodhpur\\
 Jodhpur, Rajasthan, India\\
Email: patil.3@iitj.ac.in}
}


%Telephone: (800) 555--1212, Fax: (888) 555--1212}


\maketitle

\begin{abstract}
Digital human identity management, can now be seen as a social necessity, as it is essentially required in almost every public sector such as, financial inclusions, security, banking, social networking $etc$. Hence, in today's rampantly emerging world with so many adversarial entities, relying on a single biometric trait is being too optimistic. In this paper, we have proposed a novel end-to-end, Unified Biometric ROI Segmentation Network ($UBSegNet$), for extracting region of interest from five different biometric traits $viz.$ face, iris, palm, knuckle and 4-slap fingerprint. The architecture of the proposed $UBSegNet$ consists of two stages: (i) Trait classification and (ii) Trait localization. For these stages, we have used a state of the art region based convolutional neural network (RCNN), comprising of three major parts namely convolutional layers, region proposal network (RPN) along with  classification and regression heads. The model has been evaluated over various huge publicly available biometric databases. To the best of our knowledge this is the first unified architecture proposed, segmenting multiple biometric traits. It has been tested over around $5000*5 = 25,000$ images ($5000$ images per trait) and produces very good results. Our work on unified biometric segmentation, opens up the vast opportunities in the field of multiple biometric traits based authentication systems. 
\end{abstract}


% \fi
\IEEEpeerreviewmaketitle

\section{Introduction}
In the prevailing world, biometric traits used for identification of an individual, is the best possible user friendly, effective and efficient human identity management solution. Over the last few decades, many different biometric traits have been developed and explored extensively, such as face~\cite{ajmera20143d}, iris~\cite{nigam2014iris}, fingerprint~\cite{Singh2012}, knuckle~\cite{nigam2013multimodal}, palm~\cite{badri}, ear~\cite{nigam2014robust} $etc$. Due to the uniqueness of these biometric traits they have achieved similar performance with an improved user experience, as compared to other existing token/knowledge based alternatives and people have started using them in many places regarding their security concerns. Traditionally in many places where security is concerned, password and PIN based methods had been used in the past. In password based authentication systems, every person needs to remember his or her password. It has proved to be an efficient method (only up-to few passwords) but often leads to people forgetting their passwords. Also they are pretty easy to crack, which results in huge losses to individuals in terms of privacy as well as finance. Because of these limiting nature of password related security, biometric identification and authentication methods are preferred over traditional  passwords  and  PIN  based  methods. The physical nature of the physiological biometric based identification, is the key to its secure trait, therefore it has grown to be a vast area of research in recent times.

Any biometric based authentication system contains the following six modules, (i) Data acquisition, (ii) Region of Interest (ROI) extraction, (iii) Quality estimation, (iv) Data pre-processing (v) Feature extraction and finally (vi) Matching and fusion. 

\begin{figure}[htp]
	\begin{center}
		\includegraphics[height=0.45\linewidth,width=0.95\linewidth]{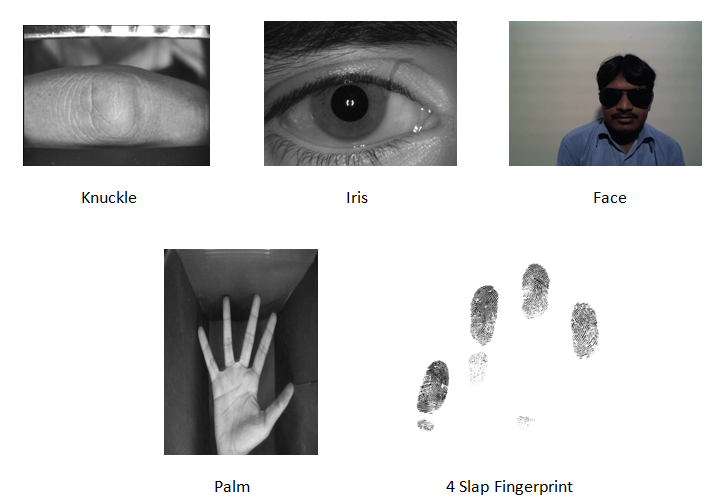}
	\end{center}
	\begin{center}
	\caption{ Examples of the Biometric Traits used}
	\label{fig:1}
	\end{center}
\end{figure}

\textbf{Motivation : } The ROI extraction is a very early step and one can easily infer that its performance plays a pivotal role in the overall system performance as all subsequent modules has to work over the region currently extracted. Several state of the art methods are available for the segmentation of any individual biometric trait. Since uni-modal systems performance got limited due to several external and environmental factors, researchers have moved to multi-modal biometric system, which uses combination of several traits for identification. Therefore a robust and efficient multi-modal biometric ROI extraction algorithm is necessarily required. Few example traits utilized are shown in Figure \ref{fig:1}.

\textbf{Related Work : }  Huge amount of work has already been done in order to segment individual biometric ROI. But to best of our knowledge no multi-modal ROI extraction has been reported till now.

In ~\cite{Singh2012}, four slap fingerprint segmentation has been performed by clustering and averaging pixel intensity of various non-overlapping boxes to localize the four fingerprints. Several iris segmentation methods has been discussed in ~\cite{bendale2012iris}. They mostly utilize circular hough transform or its variants for segmenting the inner circle. For segmenting the outer circle they applies the circular integro-differential operator. The face detection algorithm proposed by Viola and Jones~\cite{viola2004robust}, has been extensively used till date that uses a novel representation of images known as integral image and formulates a classifier using cascaded AdaBoost classifiers. Similarly, there are several knuckleprint ROI extraction algorithms such as the one presented in~\cite{Jaswal2017} and~\cite{nigam2016finger}. They have used curvature gabor filters to estimate the middle knuckle line and middle knuckle point. Finally, an area in accordance with the image size around the point has been considered. In~\cite{6306747}, palm region has been extracting using similarity constraints such as thresholding and few spatial information. 

\textbf{ Contribution : } The major contribution of our work is to provide a single fully automated network for biometric ROI extraction that can be trained and perform well for a vast variety of available biometric traits. In this paper, we have proposed an unified biometric ROI segmentation network ($UBSegNet$), which can take input as any face, iris, palm, knuckle or four slap fingerprints image and provides the actual region of interest in that image as shown in Fig. \ref{fig:7}.  Also, it can classify the extracted ROI into different biometric traits, so that it can later be matched with the appropriate gallery sample. To the best of our knowledge, this is the first unified deep learning architecture utilized to classify and localize any type of biometric sample.

\textbf{Justification for Model Selection : } We have propose a method for region extraction using deep convolutional neural network. In traditional object detection and classification approaches, first image is passed through a region proposal algorithm or network ($RPN$), that returns multiple prospective candidate bounding boxes. These boxes are later given to a classifier for classification as an input. One example of such an approach is the Region based Convolutional Neural Network (RCNN)~\cite{Girshick_2014_CVPR}, in which the region proposals are fed to a CNN classifier, but such models have very high time complexities with a  good accuracy. Object Detection has also been performed as a single regression problem, as attempted in the case of YOLO~\cite{redmon2016you}. Even though it is faster than RCNN, the issue with YOLO is that it several times it gives inaccurate results. Also its training requires more efforts and we have observed that in the case of biometric samples, its features  are not very well transferable. Hence, we have selected a faster version of $RCNN$ which gives much better accuracy as well as transferable features as compared with YOLO, named as Faster RCNN~\cite{ren2017faster}.
 
Rest of the paper has been organized as follows: Section 2, explains in detail the proposed Deep CNN architecture. Section 3, discuss the databases and testing protocol. Section 4, provides the experimental results and Section 5 concludes the paper. 

\begin{figure}[htp]
	\begin{center}
	
		\includegraphics[scale=0.5,width=1.09\linewidth]{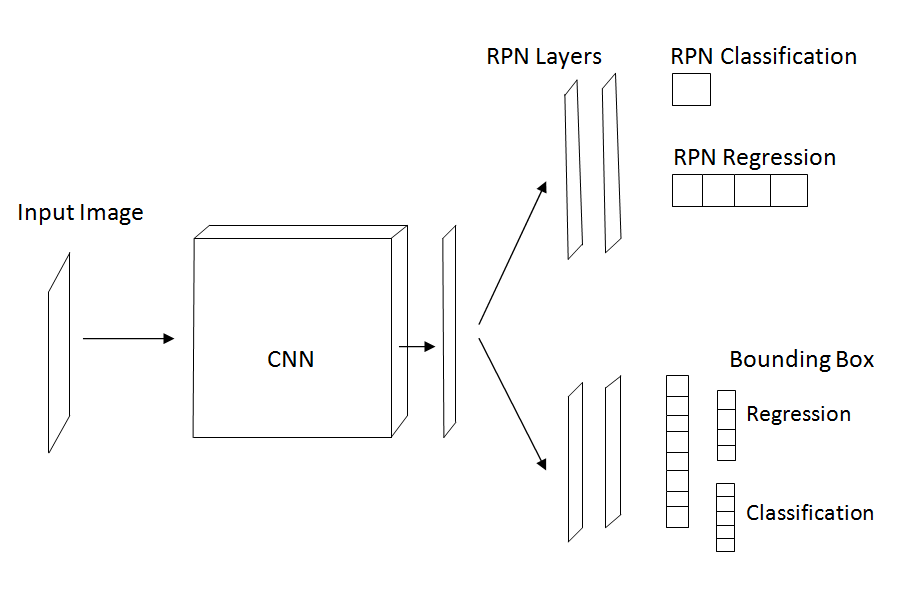}
	\end{center}
	\begin{center}
	\caption{Overview of the proposed network. There are two heads, (a) Classification and (b) Bounding Box heads.}
	\label{fig:2}
	\end{center}
\end{figure}
\section{Proposed UBSegNet}
In this section, we have discussed all the techniques that we have utilized in order to train the proposed $UBSegNet$, so as to obtain satisfactory results in the case of a multi-modal segmentation network. An overview for the proposed network has been shown in Figure \ref{fig:2}.

\subsection{UBSegNet : The Network Architecture}
The proposed $UBSegNet$, mainly consisting of these components, a) A set of shared convolutional layers for extracting the discriminative features, b) A region proposal network for regenerating candidate bounding boxes, c) ROI ( Region of Interest) Pooling layer and finally d) A regression and a classification heads as shown in Figure \ref{fig:2}.

\textbf{(a) Shared layers : } The Convolutional layers are shared between regression and classification heads and the region proposal network. Generally, these layers can be take from any popular networks (pre-trained over ImageNet) such as the ResNet or VGG/AlexNet. In the proposed network after some experimentation (over available public biometric databases), we have inferred that utilizing few layers of ResNet~\cite{he2016deep} is the best option. Such empirical analysis enable us to extract many discriminative features. Since over dataset is totally different from the one over which ResNet got initially trained, we have carefully pruned the network from the end up-to a point where one can achieve desired accuracy with lesser layers and very fast. It has been observed that a $87$ layered network (subset of full ResNet) gives similar results, for our dataset with much smaller time and space complexities.

\begin{figure}[htp]
	\begin{center}
		\includegraphics[height=0.7\linewidth,width=1.0\linewidth]{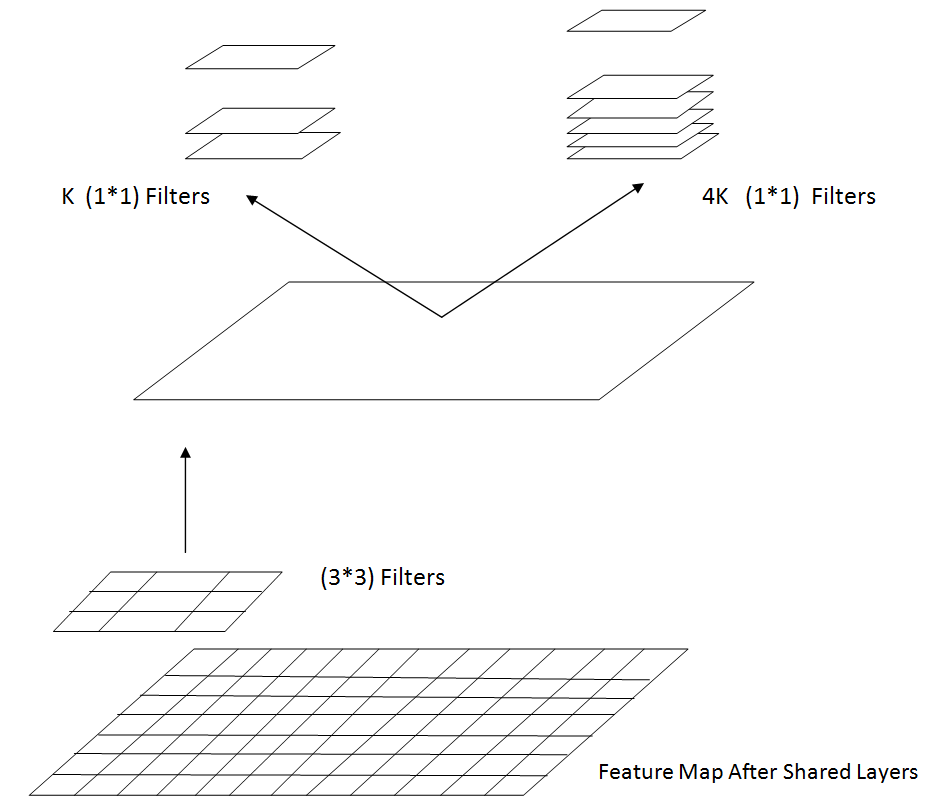}
	\end{center}
	\begin{center}
	\caption{ Region Proposal Network Implementation}
	\label{fig:3}
	\end{center}
\end{figure}

\textbf{(b) Region Proposal Network : } The Region Proposal Network (RPN) proposes various regions, over which the regression and classification heads are applied. The RPN takes a $n*n$ matrix, as an input, from the feature map obtained from previously defined shared layers. It then considers several anchor boxes of different scales and aspect ratios (as shown in Fig. \ref{fig:3}), so as to select the best fit anchor box for every ground truth bounding box. Later on, it selects the coordinates of these anchor boxes by regressing them $w.r.t$ the ground truth bounding box. The similarity between the anchor and the bounding boxes are measured using Intersection over Union ($IOU$). For each of the bounding box, at-least one anchor has to be chosen. These anchor boxes are further pruned to one (in case if they are more) per bounding box, using non-maximum suppression($NMS$). The $RPN$ network, gives an output of $4K$ and $K$ values, signifying the coordinates of the $K$ anchor boxes ($i.e.$ $4$ values per box) and the probability ($i.e.$ one value per box) of box existence respectively. 

\begin{figure}[htp]
	\begin{center}
		\includegraphics[scale=0.7,width=0.95\linewidth]{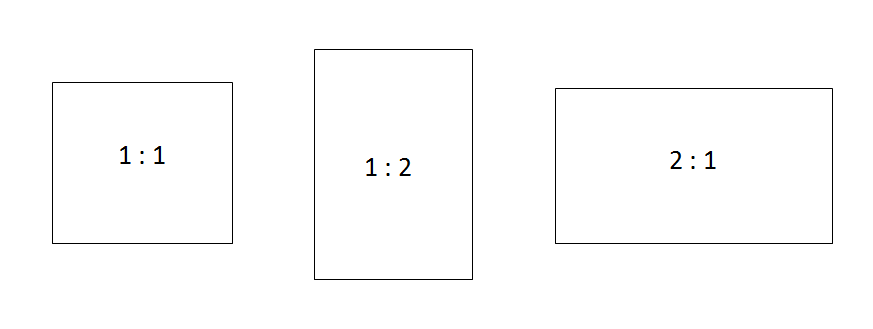}
	\end{center}
	\begin{center}
	\caption{ Anchor Boxes}
		\label{fig:4}
	\end{center}
\end{figure}

In Figure \ref{fig:3}, we have shown the implementation details of the region proposal network. It takes an input of size $3*3$ and uses a convolution layer to learn $512$ filter of size $3*3$. The $4K$ regression and $K$ probability values are computed by learning $4K$ and $K$ convolution filters of size $1*1$ of depth $512$ respectively, over the obtained $512$ feature maps. The anchor boxes are chosen to be scale and shape invariant. Three aspect ratios has been considered for anchor box at three different scales as shown in Figure \ref{fig:4}. Hence, a  total of nine anchor boxes has been considered for each $3*3$ window under consideration. 
%Three aspect ratios has been considered for anchor box at three different scales as shown in Figure \ref{fig:4}. Hence, a  total of nine anchor boxes has been considered for each $3*3$ window under consideration. 

\textbf{(c) ROI Pooling Layer : } The ROI pooling takes as input, an arbitrary sized matrix (basically a region proposed by $RPN$) and converts it to a fixed size vector ($14*14$), and apply max pooling over such a re-sized grid. It is easy to back propagate through this layer as it is just a max pooling applied over to different regions of a feature map. 

\textbf{(d) Classification and Regression heads : } Finally, the network turns into two heads predicting the class scores and bounding box coordinates. The multiple regions obtained after ROI pooling are finally fed to a network consisting of a few convolutional layers and a few fully connected layers to predict the class scores and the regression parameters (bounding boxes coordinates). The layer specific details of the proposed $UBSegNet$ has been shown in Table \ref{table:2}.

\begin{table}
\begin{tabular}{ |p{1.7cm}||p{1.1cm}|p{1.1cm}|p{.8cm}|p{1.3cm}|}
 \hline
 \textbf{Network  Component} &\multicolumn{4}{|c|}{\textbf{Layer Specifications}} \\
 \hline
 &Type, \# Filters & Filter Size & Batch Nor & Activation Function\\
 \hline
 \hline
 Shared Layers &\multicolumn{4}{|c|}{First $87$ ResNet Layers}\\
 \hline
 \hline
 \multirow{2}{*}{Region} & Conv2D, 512 & 3*3 & Yes & Relu\\\cline{2-5}
 \multirow{2}{*}{Proposal}                 & Per Anchor 4 & 1*1*512 & \multicolumn{2}{|c|}{For BB Regression}\\\cline{2-5}
 \multirow{2}{*}{Network}& Per Anchor 1 & 1*1*512 &\multicolumn{2}{|c|}{For Classification Score}\\\cline{2-5}
 \hline
 \hline
 ROI Pooling &\multicolumn{4}{|c|}{Resize ever block to 14*14 and Maxpooling}\\
 \hline
 \hline
 \multirow{2}{*}{Classification} & Conv2D, 512 & 1*1 & Yes & Relu\\\cline{2-5}
 \multirow{2}{*}{Head} & Conv2D, 512 & 3*3 & Yes & Relu\\\cline{2-5}
 \multirow{2}{*}{} & Conv2D, 2048 & 1*1 & Yes & Relu\\\cline{2-5}
 \hline
 \hline
 \multirow{2}{*}{Regression} & Conv2D, 2048 & 1*1 & Yes & Relu\\\cline{2-5}
 \hline
 \hline
 %ROI Pooling & Conv2D, 512 & 3*3 & Yes & Relu\\
 %\hline
 %Classification & Conv2D, 512 & 3*3 & Yes & Relu\\
 %\hline
 %Regression & Conv2D, 512 & 3*3 & Yes & Relu\\
 %\hline
\end{tabular}

\caption{Architecture of $UBSegNet$, as shown in Figs. \ref{fig:2}, \ref{fig:3}}
\label{table:2}
\end{table}

\subsection{Training}
 
\textbf{(a) Ground truth generation: } The ground truth with respect to face, iris, palm, knuckle and 4-slap fingerprint has been generated using \cite{viola2004robust}, \cite{bendale2012iris}, \cite{badri}, \cite{Jaswal2017}, \cite{Singh2012} respectively. The RPN, outputs 4K and K values corresponding to K anchor boxes for each $n*n$ input. We have used $IOU$ as a similarity measure between the anchor boxes and the bounding boxes provided as ground truth. The anchor boxes with the maximum IOU while compared with the ground truth are given high probabilities,  termed ``positive". It is ensured that each of the bounding boxes has to have at-least one positive anchor box corresponding to it. 

\textbf{(b) Training RPN network : } Initially, we have trained the region proposal network along with the shared layers using the above computed ground truths for the RPN. We are training $UBSegNet$ from scratch rather than considering pre-trained weights in order to make out the trained model as problem specific as possible. One has to notice, that RPN along with the shared layers has to be trained as an end-to-end network so as to achieve good performance.

\textbf{(c) Training classification and regression heads : } In the next step, we have to train the classification and the regression heads using the obtained region proposals. This also has to be carried out in end-to-end fashion through ROI pooling layer and shared convolutional layers. 

\textbf{(d) Fine Tuning RPN : } Once we have trained the shared layers  for RPN and both heads (as in Steps (b),(c)), the best possible and discriminative features have been learned at shared layers attaining the maximum accuracy. But the problem is, that RPN is trained as end-to-end in Step (b), along with shared layers. Hence, we have fine tuned the RPN layers keeping the shared layers frozen, in order to learn the anchor box prediction and their probabilities. 

\textbf{(e) Fine tuning the Classification and Regression heads : } Similarly, the classification and regression heads has to be fine tuned in order to take a different feature map as an input, keeping the weights of shared layers frozen, to get satisfactory results.

\textbf{(f) Losses : } In order to train such a deep network, we have considered four kinds of losses, (i) RPN regression loss, (ii) RPN classification loss, (iii) Final regression loss and (iv) final classification loss. During every epoch, it first trains the RPN network followed by training the final regression and classification heads. The losses used for RPN classification as well as trait classification are ``binary cross-entropy" and ``categorical cross-entropy" respectively. The mean squared error (MSE), loss function has been used for regression of both region proposal network (RPN) as well as bounding boxes. 

\textbf{Size invariant network : }  Our network can take input of any size (size invariant), mainly because of this tweaked implementation of region proposal network (RPN) and the classification and regression heads. The region proposal network has been implemented as convolutional layers and the ground truth is corresponding to the image size, making our RPN size invariant. In the case of classification and regression heads, the ROI pooling layer serves this purpose, as the pooling layers take in any arbitrary sized region of interest (ROI) and pools it into a fixed sized output as discussed above. This fixed sized output has been fed to a network consisting of convolutional and fully connected layers, making the classification and regression heads size invariant too.

All the network hyper-parameters, have been selected ``emperically'' by maximizing the system performance over a validation set.

\section{ Testing Strategy Used}
 
We have used over $10,000$ images for training while around $5000$ images has been used for testing, per trait, in-order to generate the proposed $UBSegNet$. The trained $UBSegNet$ has been tested by evaluating intersection over union ($IOU$), that we have used to obtain the accuracy of our proposed network. It is the most widely used evaluating parameter, to check the efficiency of any algorithm/network, for object localization. Iterative thresholding has been applied to over each of the traits individually, as well over all the traits to determine the individual trait as well as  overall performance analysis.  

\begin{figure}[htp]
	\begin{center}
		\includegraphics[height=1.09\linewidth,width=1.1\linewidth]{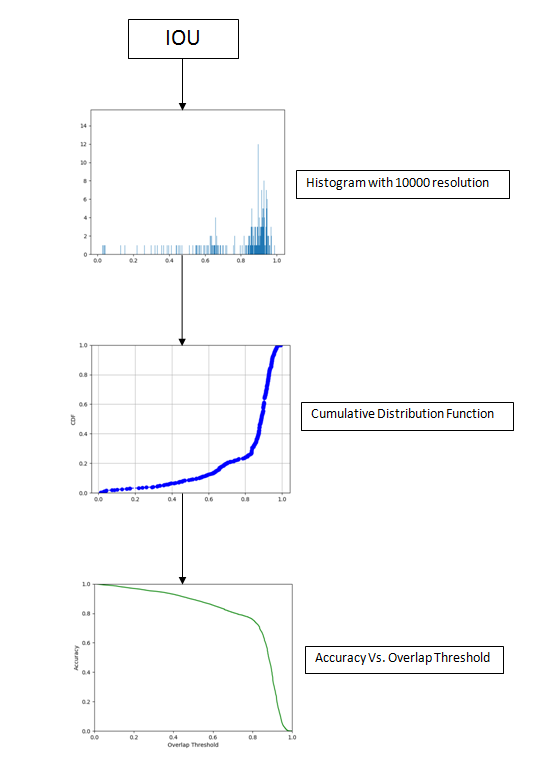}
	\end{center}
	\begin{center}
	\caption{Steps involved for generating Accuracy Vs IOU Graph}
		\label{fig:5}
	\end{center}
\end{figure}

\textbf{(a) Accuracy Vs IOU Graph : } To visualize system performance, we have plotted a graph, showing accuracy at each threshold for each trait as well as overall, as shown in Fig. ~\ref{fig:5}. The IOU ranges from $0$ to $1$. Where $0$, indicates that the boxes do not match at all and $1$ indicates that the boxes are perfectly matched. When the threshold is high the number of images (in \%) having IOU more than the threshold will be less, where as it is $100\%$ at $0$ threshold. 

Such a graph can be plotted as follows: Compute IOU of predicted and ground truth boxes. The predicted boxes having the same predicted predicted as that of the ground truth along with their respective distance from ground truth has been considered to match the boxes in images containing multiple boxes ($e.g.$ four slap images). Generate a histogram over IOU values at an step of $0.00001$ so as to get a smooth curve. Normalize it, so as to get a probability distribution function (PDF) and compute its cumulative distribution function (CDF). 

The accuracy at each IOU threshold ($i_t$) can be defined as : $Accuracy = \frac{\#\ test\ images\ with\ IOU\ >= i_t}{\#\ test\ images}$ and can be computed using the Eq. \eqref{eq:11}, as shown in Fig. \ref{fig:5}.
    
    %\begin{equation}
    %    Accuracy = \frac{\#\ test\ images\ with\ IOU\ >=\ IOU\ threshold}{\#\ test\ images}
    %\end{equation}
    
    \begin{equation}
        Accuracy=1\ -\ cdf\ +\ Value\ of\ histogram
        \label{eq:11}
    \end{equation}
    
    %and then we plot accuracy vs. overlap threshold graph to visualise the performance of our network.

%The steps used to compute the graph are:-

%\begin{itemize}
   % \item First the intersection over union of the predicted boxes with respect to the ground truth boxes were calculated. The predicted boxes having the same class predicted as that of the ground truth and their respective distance from the ground truth boxes were considered to match the boxes in images containing multiple boxes. The rest of the bounding boxes were considered to give 0 intersection over union values. 
    %\item Next a histogram was plotted using the intersection over union values obtained using the above step, the resolution of the histogram was taken to be $10000$ so as to form a curve as smooth as possible.
    %\item After obtaining the histogram, the histogram is normalized and then a probability distribution function is formulated from it.
    %\item The probability distribution function is then converted into cumulative distribution function, by summation of all the previous probability values at a particular point.   
    %\item We compute accuracy for each value of the overlap threshold ( IOU threshold ) we compute the number of images having IOU greater than or equal to that value by computing 
    
    %\begin{equation}
     %   accuracy=1\ -\ cdf\ +\ Value\ of\ histogram
    %\end{equation}
    
    %and then we plot accuracy vs. overlap threshold graph to visualise the performance of our network.
%\end{itemize}

%An overview of the proposed testing strategy is shown in Figure \ref{fig:5}.

\textbf{(b) Precision and Recall : } In addition to the accuracy values, precision and recall has also been calculated for the proposed network validation as defined in Eqs. \eqref{eq:p}, \eqref{eq:r}. 

\begin{equation}
Precision = \frac{\#\ of\ correct\ boxes\ predicted} {Total\ No.\ of\ boxes\ predicted}
\label{eq:p}
\end{equation}

\begin{equation}
Recall = \frac{\#\ of\ correct\ boxes\ predicted} {Total\ No.\ of\ Ground\ truth\ boxes}
\label{eq:r}
\end{equation}

Precision and recall are calculated so as to validate our approach, while calculating accuracy we only consider the true predicted boxes and not all the predicted boxes, hence the need for precision. Similarly the intersection over union values calculated are with respect to the ground truth bounding boxes, but it may so happen that all the ground truth boxes are not considered while calculating accuracy, therefore we take into account this detail while computing the recall values.

\begin{table*}
\begin{tabular}{ |p{2.8cm}||p{1.15cm}|p{1.15cm}|p{1.15cm}| |p{1.15cm}|p{1.15cm}|p{1.15cm}||p{1.15cm}|p{1.15cm}|p{1.15cm}| }
 \hline
 \textbf{Biometric Traits} &\multicolumn{3}{|c||}{\textbf{Accuracy} }&\multicolumn{3}{|c||}{\textbf{Precision}}&\multicolumn{3}{|c|}{\textbf{Recall}} \\
 \hline
  &\multicolumn{3}{|c||}{Overlap IOU Threshold }&\multicolumn{3}{|c||}{Overlap IOU Threshold }&\multicolumn{3}{|c|}{Overlap IOU Threshold }\\
 \hline
     & \textbf{0.35} &\textbf{0.5}&\textbf{0.65}& \textbf{0.35} &\textbf{0.5}&\textbf{0.65}& \textbf{0.35} &\textbf{0.5}&\textbf{0.65}\\
 \hline
 \textbf{Knuckle}   & 99.14    &93.72&   78.3 &98.94&93.48&77.97&98.77&93.38&77.00\\
 \hline
 \textbf{Iris}&   96.78  & 90.24   &87.42&94.84&88.43&85.67&   96.78  & 90.24   &87.42\\
 \hline
 \textbf{Face} &98.46 & 98.34& 98.27 &97.46&97.34&97.27&97.91&97.77&97.70\\
 \hline
 \textbf{Palm}    &99.66 & 99.52&  98.33&98.14&98.00&96.83&97.80&97.66&96.50\\
 \hline
 \textbf{4 Slap Fingerprint}&   99.74  & 96.94&84.38&99.24&96.46&83.95&98.70&95.93&83.49\\
 \hline
 \textbf{All Traits}& 99.00  & 96.62   &90.53&98.51&96.14&90.08&97.97&95.61&89.58\\
 
 \hline
\end{tabular}
\caption{The Accuracy, Precision and Recall Values at different Overlap (IOU) thresholds}
\label{table:1}
\end{table*}

\textbf{(c) Four Slap Fingerprint Testing : } Testing for all the other traits has been performed by taking the bounding box predicted with the maximum probability and finding the IOU for it, $w.r.t$ the ground truth. But for 4-slap fingerprint, the testing mechanism has to be varied as there are multiple boxes in the ground truth. Hence, we have used a method using $x$-axis projection and distances to address this issue.

We first take the projection of the bounding boxes on $x$-axis, and compute the overlap between each of the predicted bounding boxes. If the overlap between any two boxes, is greater than a empirically selected threshold, then drop those boxes which are below to any other reported block. This is done as multiple fingerprint bounding boxes have been reported by $UBSegNet$, but we know that there must not be two fingerprint one below other. Basically we have dropped the lower fingerprint in these cases. After getting all predicted bounding boxes, we have computed the distance each of them $w.r.t$ the ground truth boxes and then considering the box with minimum distance as the ``corresponding'' box. Finally, we have computed the IOU between, ``corresponding'' predicted and ground truth boxes.

%During observation of the 4 slap fingerprint images, we found that most of the region of interests are in the upper part of the image so we eliminate the box which has lower vertical coordinate values.

%Now the problem arises that which predicted box must be considered for finding the intersection over union value with respect to each of the ground truth boxes. This issue is resolved by considering the distance of the predicted boxes with respect to the ground truth boxes and then considering the box with minimum distance for calculating the IOU.

\section{Experimental Results and Discussion}
\begin{figure}[htp]
	\begin{center}
		\includegraphics[height=0.8\linewidth,width=1.09\linewidth]{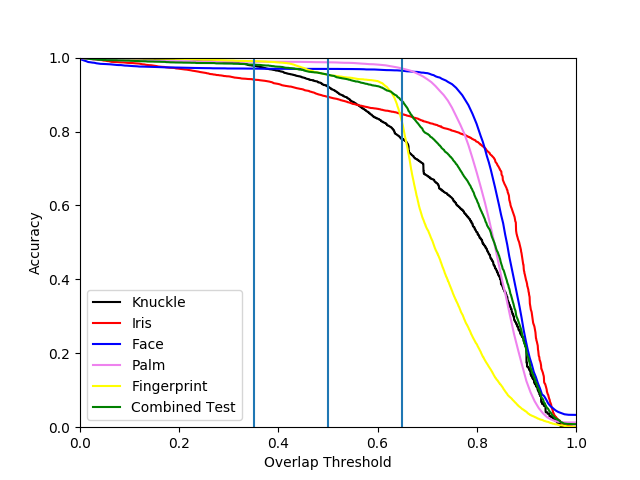}
	\end{center}
	\begin{center}
	\caption{ Accuracy Vs. Overlap Threshold Graph when tested for all the traits individually and combined}
		\label{fig:6}
	\end{center}
\end{figure}

The experiments in accordance with the testing strategy and parameters as mentioned above has been conducted in two fold, (i) Individual biometric traits in which $5000$ images has been used for testing. The Accuracy vs. IOU threshold graph is shown in Fig. \ref{fig:6}, where different colours have been used to plot the graphs for different traits. 
%The traits and the colours used for them are : Knuckle - Black, Iris - Red, Face - Blue, Palm - Violet, 4 Slap Fingerprint - Red and the graph when tested with a test set having all the traits is depicted in green colour.
In Fig. \ref{fig:6}, the combined Accuracy Vs. Overlap Threshold graph validates the effectiveness of our proposed $UBSegNet$. Table \ref{table:1}, shows the values obtained for Accuracy, Precision and Recall for the all experiments performed. One can observe that the network produces high accuracy even up-to $0.4\ to\ 0.5$ overlap IOU threshold for almost all the traits. Slight accuracy drop has been observed when overlap IOU threshold becomes more than $0.6$, especially for knuckle. It may be as the curvature like features are evenly distributed and most of the previous approaches~\cite{nigam2016finger} tried to obtain the centre line or the center point. Network may not be able to capture such a symmetry. Even though it is performing better than the existing systems. 

\begin{figure*}[htp]
	\begin{center}
		\includegraphics[height=0.27\linewidth,width=1.0\linewidth]{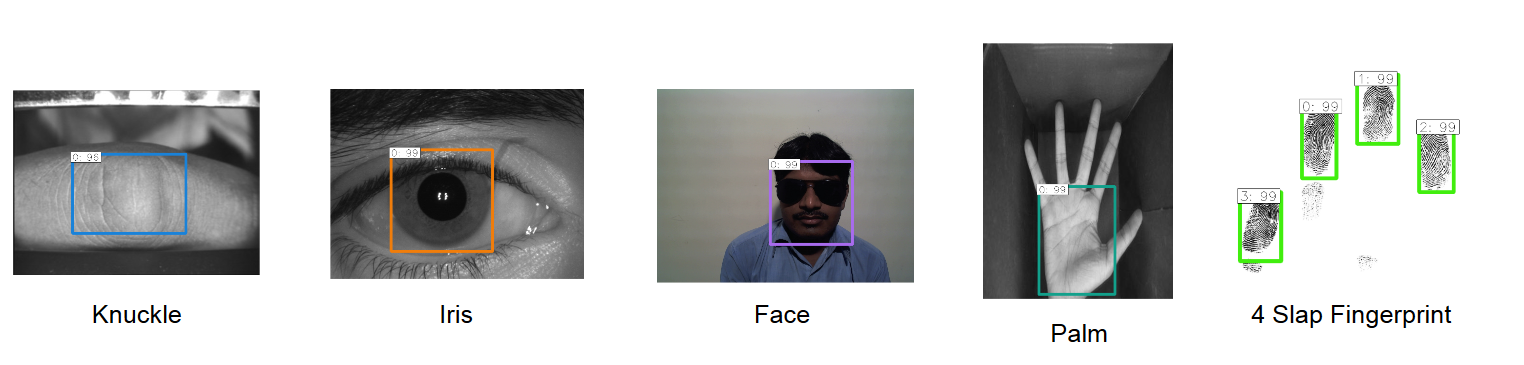}
	\end{center}
	\begin{center}
	\caption{ROI obtained for various traits using UBSegNet. Colored box shows our predicted box along with its probability}
		\label{fig:7}
	\end{center}
\end{figure*}

Network was performing excellently for face and palm traits. The prime reason behind this is that these traits contain features that are easily distinguishable from the others under their respective region of interests. In terms of precision and recall very similar trends have been observed. From Table \ref{table:1}, Fig. \ref{fig:6} one can infer that the proposed network has been performing very well across all the traits. Some network predictions are shown in Figure \ref{fig:7}.

 \textbf{Comparative Performance Analysis : } To best of our knowledge, this is the first ever proposed multi-modal biometrics deep learning segmentation network. Hence, we have not compared our results with any other method. Although, one can compare it with the existing techniques, such as \cite{viola2004robust}, \cite{bendale2012iris}, \cite{badri}, \cite{Jaswal2017}, \cite{Singh2012}, but such comparison may not be justified due to two reasons : (i) They have been tested only over single trait (we have performed multi-class classification) and (ii) None of them have used deep learning. Still we have observed that the proposed network performs better that previous individual trait techniques.
 
\textbf{Network Limitations : }  We have tried to train $UBSegNet$ on ear data. It became a very challenging task due to these three reasons : (1) Less amount to annotated data,  (2) unavailability of any good segmentation algorithm, (3) It has features very similar face and started our network started to confuse it with face. Due to its same texture as that of face and similar shape the region proposal was quite effective but the classification accuracy was not satisfactory. Hence we inferred that, if two traits are ``similar'' or ``subset'' of each other, network will not be trained optimally. In future, we will incorporate ear as well, extending it to a six class biometric segmentation problem.

\section{Conclusion}
 In this paper, we have proposed a novel end-to-end network for extracting ROI from any biometric trait (face, iris, palm, knuckle and 4-slap fingerprint). To best of our knowledge this is the first unified architecture proposed, segmenting multiple biometric traits. It has been tested over around $5000*5 = 25,000$ images and produces very good results. Our work on unified biometric segmentation, opens up the vast opportunities in the field of multi-biometric authentication systems. 
%In the case of knuckle as you can see in Figure \ref{fig:1}, the curvature like features are evenly distributed, therefore most of the previous approaches~\cite{nigam2016finger} try to obtain the centre line or the center point, most of our test results are satisfactory as these contain the centre line but the sizes of the region of interest are smaller in testing resulting in smaller IOU values.  

%\section*{Acknowledgment}
%more thanks here
%\begin{thebibliography}{1}
  %0.5em minus 0.4em\relax Harlow, England: Addison-Wesley, 1999.
%\end{thebibliography}
\bibliographystyle{IEEEtran}
\bibliography{IEEE_bib}
\end{document}